\documentclass[10pt,twocolumn,letterpaper]{article}

\usepackage{iccv}
\usepackage{times}
\usepackage{epsfig}
\usepackage{graphicx}
\usepackage{amsmath}
\usepackage{amssymb}

\usepackage[breaklinks=true,bookmarks=false]{hyperref}

\iccvfinalcopy % *** Uncomment this line for the final submission

\def\iccvPaperID{****} % *** Enter the ICCV Paper ID here

\ificcvfinal\fi

\begin{document}

%%%%%%%%% TITLE
\title{3rd Place Solution to ICCV LargeFineFoodAI Retrieval}
% \title{3st Place Solution to Google Landmark Retrieval 2021}

\author{
Yang Zhong\\
East China Normal University\\
{\tt\small zhongyangtony@163.com}
\and
Zhiming Wang\\
Tsinghua University\\
{\tt\small wang-zm18@mails.tsinghua.edu.cn}
\and
Zhaoyang Li\\
OPPO Research Institute\\
{\tt\small zhaoyangli1996@gmail.com}
\and
Jinyu Ma\\
OPPO Research Institute\\
{\tt\small majinyu1@oppo.com}
\and
Xiang Li\\
OPPO Research Institute\\
{\tt\small lixiang8@oppo.com}
}

\maketitle
\ificcvfinal\thispagestyle{empty}\fi

%%%%%%%%% ABSTRACT

%%%%%%%%% ABSTRACT
\begin{abstract}
This paper introduces the 3rd place solution to the ICCV LargeFineFoodAI Retrieval Competition on Kaggle. Four basic models are independently trained with the weighted sum of ArcFace  and Circle loss, then TTA and Ensemble are successively applied to improve feature representation ability. In addition, a new reranking method for retrieval is proposed based on diffusion and k-reciprocal reranking.
Finally, our method scored 0.81219 and 0.81191 mAP@100 on the public and private leaderboard, respectively.
\end{abstract}
\section{Introduction}
ICCV LargeFineFoodAI Retrieval is the first fine-grained food retrieval competition on Kaggle. The purpose of this competition is to find the most similar gallery images for each query image. As a fine-grained visual analysis task, some of images are hard to be manually discriminated, which increases the difficulty of retrieval.

The LargeFineFoodAI Retrieval Track shares the same training dataset with LargeFineFoodAI Recognition Track, which consists of 317,277 images and covers 1000 food classes. Specific to retrieval track,  images from extra 500 food classes form  the testing dataset, which consists of 10,000 query images and 209,562 gallery images. Since there are no overlapped food classes between training and testing dataset, to some extent, it makes challenges to the transfer ability of trained models. 

The rest of this paper is organized as follows. Section \ref{Traning} describes backbones and training strategies.  Section \ref{post} introduces the ensemble and post processings in detail. Lastly, the summary is made in Section \ref{summary}.

\begin{figure*}[!htbp]
	\centering
	\includegraphics[width=\linewidth]{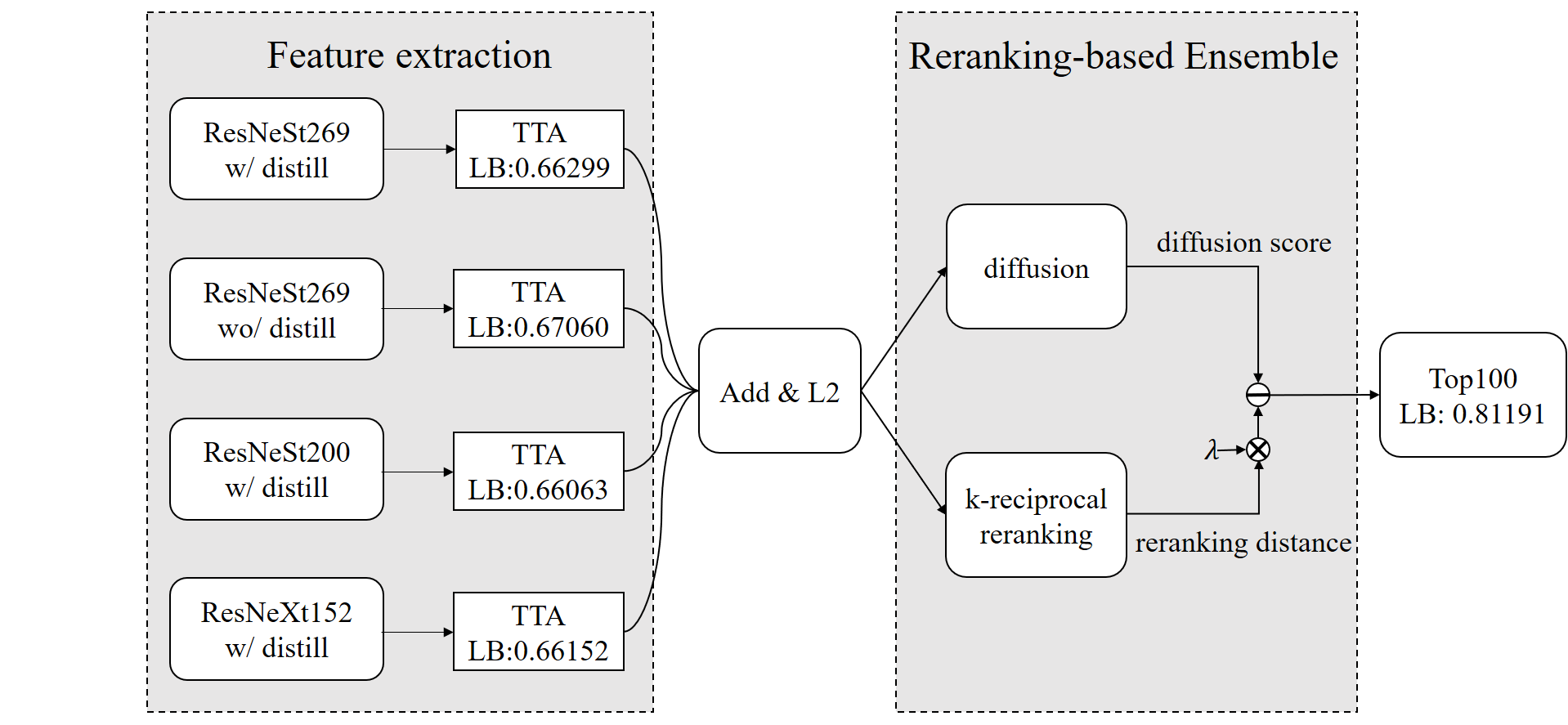}
	\caption{Architecture of our post processing for retrieval, where ``w/" means ``with", ``w/o" means ``without".}
	\label{arch}
\end{figure*}

\section{Training Strategy}
\label{Traning}
To obtain food image features, convolutional neural networks are employed through our pipeline. The backbones we used are ResNeXt152 \cite{xie2017aggregated}, ResNeSt269 \cite{zhang2020resnest}, and ResNeSt200 \cite{zhang2020resnest}. We train all the models listed above with our training settings and distill all of them. Then we use test-time augmentation (TTA) to improve their performance.

\subsection{loss Function}
To improve the performance, we tried several loss functions, including Cross Entropy loss, ArcFace loss \cite{deng2019arcface} and Circle loss \cite{sun2020circle}. Arcface performs better than the others, as it can enhance the discriminative power of fine-grained classification models. Then we tried the combination of different loss functions and found that adding Circle loss to ArcFace with a proper weight can accelerate convergence and further improve the performance of the model. 

We use ArcFace loss \cite{deng2019arcface} with a margin of 0.2 and a scale of 32 to train the models, then we use the Circle loss \cite{sun2020circle} with m of 0.25 and delta of 32. In order to balance the role of the two functions in training, our loss is computed as: 
\begin{equation}\label{our_loss}
 \mathcal{L} = \delta_{0}\mathcal{L}_{a} + \delta_{1}\mathcal{L}_{c}
\end{equation}
where $\mathcal{L}$ is the loss function we finally use. $\mathcal{L}_{a}$ denotes ArcFace loss and $\mathcal{L}_{c}$ denotes Circle loss. $\delta_{0}, \delta_{1}$ are the weights of two loss functions, we set $\delta_{0}$ as 1 and $\delta_{1}$ as 1/$\beta$, where $\beta$ is the batch size used for training. During training, ArcFace loss plays a major role in the early stage and Circle loss works during the late stage.

\subsection{Backbone}
Eight NVIDIA Tesla-V100 GPUs are used for training. During the training stage, we apply augmentations including random scale and center crop, horizontal flip with the probability of 0.5, 15-degree rotation, and color jitter \cite{krizhevsky2012imagenet} and set the training image size to 512x512. For ResNest269, we set the training image to 416x416, because our research shows that smaller image size is better for retrieval, but due to the limit of time and computation resources, we can only change one model. And the loss function we use is loss function(1).

We use different learning rates for different models in order to fit their training batch sizes and maximize the utility of GPUs. Our optimizer use SGD with momentum of 0.9. For the learning rate scheduler, cosine annealing\cite{loshchilov2016sgdr} is used and we set T as the max epoch in training. For retrieval, we output the features after the final GAP layer and normalize them, the dimension of that is 1x2048 per image. The early epoch of ResNest269 is used in further ensemble because our research shows that the choice of epoch will remarkably influence the result, as shown in Table\ref{tab:results}.

\begin{table}[htbp]
\begin{center}
\setlength{\tabcolsep}{1.8mm}{
\begin{tabular}{c c c c}
epoch & recogination acc & loss value & retrieval score\\
\hline
10 & 0.845 & 1.9585 & 0.59204\\
15 & 0.8831 & 1.7001 & 0.6473\\
20 & 0.8944 & 1.4986 & 0.64581\\
24 & 0.8992 & 1.362 & 0.64034\\
30 & 0.9031 & 1.1578 & 0.62799\\
40 & 0.9061 & 0.7997 & 0.59956\\
\end{tabular}}
\end{center}
\caption{Results of different epochs in ResNeSt269.}
\label{tab:results}
\end{table}

\subsection{Model Distillation}
We tried the traditional distillation method \cite{hinton2015distilling} with KDloss and cross entropy loss of equal weights and gained improvements. During distillation, KDloss is used to compute the loss of soft label, which is the 1000-dimension output embeddings of the teacher model and Cross Entropy loss is used to compute the loss of hard label, i.e., the ground truth. Considering the scale of our dataset and the way we distill, we set the temperature of KDloss to 3 and the weight of two loss functions to 1:1. Finally, ResNeSt269, ResNeSt200 and ResNeXt152 are used in further ensemble.

\section{Ensemble And Post Processing}
\label{post}
The architecture of post processing is shown as Figure \ref{arch}, which consists of TTA, model ensemble and reranking. Each part is introduced in detail as follows.
\subsection{TTA}

 To make the extracted features more robust, test time augmentation (TTA) trick is applied here. In practice, we utilize four kinds of test-time augmentations per test image, which are five crops, resize, horizontal flip, and rotation. This method can make up for the lack of information compared to a single input. We found that a slightly bigger input size leads to better performance, thus we use different input sizes during training and testing.  Experiments shows that TTA brings about 2 $\sim$ 4 percent gains. 
 
\subsection{Ensemble}
For model ensemble, we select 4 pretrained models which scores above 0.66 by virtue of nearest neighbor search after TTA operations. For simplicity, we ensemble these features by add operation, then L2 normalization is applied along the feature dimension. Experiments show that feature ensemble will bring  about 1\% $\sim$ 2\% improvement.

\subsection{Reranking}
To boost the final retrieval performance, here we propose a new retrieval reranking  method, which is based on  two classical retrieval methods: Yang's diffusion \cite{yang2019efficient} and k-reciprocal reranking \cite{zhong2017re}.
In the first step, diffusion and k-reciprocal reranking algorithms are conducted independently, which both take ensembled features with length 2048 as input . Diffusion algorithm outputs a similarity matrix $S$ with shape $10000 \times 209562$, whose elements reflect the similarity between each query and gallery image pair. Similarly, k-reciprocal reranking algorithm outputs a distance matrix $D$ with the same shape, whose elements express the distance between each query and gallery image pair. Given $S$ and $D$, the final similarity score $S_{final}$ is defined as follows:
	\begin{equation}
		\begin{aligned}
			S_{final} = S - \lambda \times D
		\end{aligned}
		\label{formula}
	\end{equation}
Ultimately, we sort on $S_{final}$ along the feature axis in descending order, and choose Top100 for each query as the final retrieval results. 

In practice, for diffusion algorithm, we set subgraph scale $n_{trunc}=10000$, and choose $kd=70$ nearest neighbors as the initial neighbors for each node. For k-reciprocal reranking algorithm, we set reciprocal nearest neighbors $k_1=260$,  local query expansion parameter $k_2=30$, weighting coefficient for balancing Euclidean distance and Jaccard distance $\lambda_{value}=0.3$. The improvements made by diffusion and k-reciprocal reranking on single model are about 8\% and 6\%, respectively. Limited by the number of submissions,  we did not test the performance of above two algorithms on final ensembled features. But our former experiments reveal that the more robust input features are,  the higher gains diffusion algorithm will bring. When combining two algorithms, for simplicity, we set $\lambda=1$ in Equation (\ref{formula}). Also limited by the number of submissions, we did not finetune this hyperparameter later. Previous 3 models' ensemble experiments show that our combined method brings about 0.33\% improvement than the best single retrieval algorithm.

In addition, there are also several post processing methods we have tried before. We used principal component analysis (PCA) \cite{sirovich1987low} to reduce the dimension of the features and improve the performance by about 0.4\% for the single model, but the performance even deteriorates when further combined with whitening. We also used database-side feature augmentation (DBA) \cite{arandjelovic2012three} and average query expansion (AQE) \cite{chum2007total} methods for post-processing. While DBA can improve performance to some extent, further combined with AQE results in performance degradation. Moreover, these methods can only yield limited improvement, even less than the performance of using diffusion alone. In addition, since diffusion algorithm requires that different dimensions of features are irrelevant, these methods usually did not work with diffusion. 

\section{Summary}
\label{summary}
In this paper, 3rd place solution to LargeFineFoodAI 
Retrieval is introduced in detail. The solution ensembled 4 models pretrained on the training dataset with ArcFace  and Circle loss. During the post processing phase, TTA and ensemble are applied to make features more robust, and diffusion and k-reciprocal reranking are conducted independently, then  diffusion score and reranking distance are combined to make  the final retrieval result. 
{\small
\bibliographystyle{ieee_fullname}
\bibliography{egbib}
}

\end{document}